\documentclass[10pt,twocolumn,letterpaper]{article}

\usepackage[accsupp]{axessibility}
\usepackage[applications]{wacv}
\usepackage[table]{xcolor}

\usepackage{url}            %
\usepackage{booktabs}       %
\usepackage{graphicx}
\usepackage{amsfonts}       %
\usepackage{nicefrac}       %
\usepackage{microtype}      %
\usepackage{xcolor}         %
\usepackage{algorithm}
\usepackage{tabularx}
\usepackage{graphicx}	
\usepackage{amsmath}	
\usepackage{amssymb}	
\usepackage{booktabs}
\usepackage{times}
\usepackage{microtype}
\usepackage{epsfig}
\usepackage{caption}
\usepackage{float}
\usepackage{placeins}
\usepackage{color, colortbl}
\usepackage{xcolor}
\usepackage{stfloats}
\usepackage{enumitem}
\usepackage{tabularx}
\usepackage{xstring}
\usepackage{multirow}
\usepackage{xspace}
\usepackage{url}
\usepackage{subcaption}

\definecolor{wacvblue}{rgb}{0.21,0.49,0.74}
\usepackage[pagebackref,breaklinks,colorlinks,allcolors=wacvblue]{hyperref}

\def\wacvPaperID{327} 
\def\confName{WACV}
\def\confYear{2026}

\title{PointNet4D: A Lightweight 4D Point Cloud Video Backbone for Online and Offline Perception in Robotic Applications}

\author{
Yunze Liu\textsuperscript{1} \quad
Zifan Wang\textsuperscript{1} \quad
Peiran Wu\textsuperscript{2} \quad
Jiayang Ao\textsuperscript{3} \\
\textsuperscript{1}Tsinghua University \quad\quad\quad\textsuperscript{2}University of Bristol \quad\quad\quad\textsuperscript{3}The University of Melbourne\\
}

\begin{document}

\twocolumn[{%
\maketitle
\vspace{-1.5em}
\begin{center}
\includegraphics[width=\textwidth]{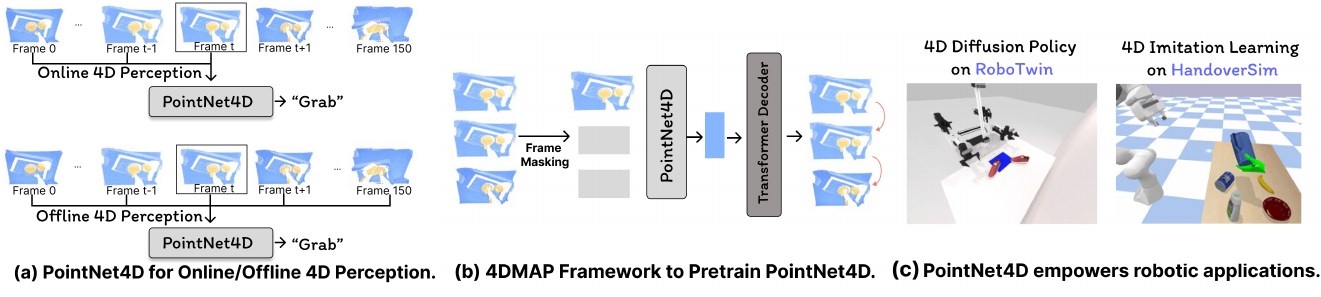}
\captionof{figure}{(a) We propose PointNet4D, a lightweight and unified network that can simultaneously handle both online real-time 
and offline perception tasks. (b) We introduce the 4DMAP pre-training method for PointNet4D to fully harness its potential. (c) We propose the 4D Diffusion Policy and 4D Imitation Learning based on PointNet4D to enhance the robot's perception capabilities.}
\label{fig:teaser}
\end{center}
\vspace{1em}
}]

\maketitle

\begin{abstract}
Understanding dynamic 4D environments—3D space evolving over time—is critical for robotic and interactive systems. These applications demand systems that can process streaming point cloud video in real-time, often under resource constraints, while also benefiting from past and present observations when available.  However, current 4D backbone networks rely heavily on spatiotemporal convolutions and Transformers, which are often computationally intensive and poorly suited to real-time applications.
We propose \textbf{PointNet4D}, a lightweight 4D backbone optimized for both online and offline settings. At its core is a \textbf{Hybrid Mamba-Transformer temporal fusion block}, which integrates the efficient state-space modeling of Mamba and the bidirectional modeling power of Transformers. This enables PointNet4D to handle variable-length online sequences efficiently across different deployment scenarios.
To enhance temporal understanding, we introduce \textbf{4DMAP}, a frame-wise masked auto-regressive pretraining strategy that captures motion cues across frames. Our extensive evaluations across 9 tasks on 7 datasets, demonstrating consistent improvements across diverse domains. We further demonstrate PointNet4D's utility by building two robotic application systems: \textbf{4D Diffusion Policy (DP4)} and \textbf{4D Imitation Learning (4DIL)}, achieving substantial gains on the RoboTwin and HandoverSim benchmarks. 
%
Code and checkpoints available: \href{https://github.com/yunzeliu/MAP}{https://github.com/yunzeliu/MAP}
\end{abstract}

\section{Introduction}

Recent advances~\cite{liu2024mamba4d,liu2023leaf,wen2022point} in video understanding make real-time 4D point cloud perception (i.e., 3D geometry over time) increasingly promising for applications in robotics, augmented reality/virtual reality (AR/VR), and embodied artificial intelligence (AI). These domains operate in a world that is inherently 4D, combining 3D spatial observations with a 1D temporal dimension. To function effectively, AI agents must interpret dynamic 4D information rather than rely on static, single-frame inputs. In AR/VR and embodied AI, agents receive continuous, variable-length streams of observations accumulating over time. This requires solving online perception problems under real-time constraints to guide immediate decisions. In contrast, in settings such as retrospective analysis, agents must take advantage of the complete fixed-length sequence for offline perception.

Despite the growing need for dynamic scene understanding, most current deployed systems still rely on static 3D backbones such as PointNet++~\cite{qi2017pointnet++}, which offer efficiency but cannot model temporal dynamics. While recent 4D backbone networks~\cite{fan2021point,wen2022point,zhang2023dpmix,liu2023leaf} based on spatiotemporal convolutions and Transformers improve accuracy, they remain too resource-intensive for online use, and often fail to generalize across variable-length input streams.

In this paper, we address these limitations by introducing \textbf{PointNet4D}, a lightweight and unified 4D backbone designed for both \textbf{online real-time} and \textbf{offline full-sequence} processing. Our approach is motivated by practical demands in applications such as robot control, where agents must act based only on current and past observations, and offline retrospective analysis, where the full sequence is available. 

To achieve this, PointNet4D introduces a \textbf{hybrid Mamba-Transformer temporal fusion block}, combining Mamba’s~\cite{gu2023mamba} efficient scanning mechanism for unidirectional temporal modeling with the bidirectional reasoning strengths of Transformers~\cite{liu2021swin}. This design is motivated by the complementary strengths of each component: Mamba excels in online settings with its state-updating and selective memory mechanisms, while Transformers are better suited for offline tasks due to their global bidirectional attention. By integrating both within a single framework, PointNet4D achieves robust performance across varied temporal conditions without sacrificing efficiency. We also develop an effective pretraining strategy, \textbf{4DMAP}, which applies masked auto-regressive learning to teach the network both spatial geometry and motion-aware features.

We evaluate PointNet4D's performance across 9 tasks on 7 datasets. More importantly, we demonstrate real-world value by deploying PointNet4D in robotics applications. On the RoboTwin benchmark~\cite{mu2024robotwin}, our 4D Diffusion Policy (DP4) achieves substantial performance gains over prior methods~\cite{ze20243d, chi2023diffusion} across 14 manipulation tasks, with only 0.3\% additional parameters and minimal computational overhead. On the HandoverSim benchmark~\cite{chao2022handoversim}, our 4D Imitation Learning (4DIL) model outperforms sequential models even in simultaneous settings, a key requirement for smooth human-robot collaboration.

By bridging online and offline temporal modeling in a lightweight design, PointNet4D offers a practical path forward for 4D perception in resource-constrained and interactive applications. Our contributions are fourfold as follows:
\begin{enumerate}
    \item We propose PointNet4D, a lightweight 4D backbone network capable of supporting diverse tasks in both online real-time and offline settings. 
    \item We introduce 4DMAP, a pretraining strategy that fully leverages the hybrid architecture of PointNet4D.
    \item We demonstrate PointNet4D’s great application potential in robotics, presenting DP4 and 4DIL, which achieve substantial performance gains on the RoboTwin and HandoverSim benchmarks.
    \item We comprehensively validate PointNet4D’s potential as a universal 4D backbone across 9 different tasks.
\end{enumerate}
\section{Related Work}
\label{sec:related}
\textbf{4D Point Cloud Video Perception.}
Point cloud videos have garnered increasing attention in recent research due to its relevance in dynamic scene analysis for robotics, AR/VR, and autonomous systems. Many recent works have focused on modeling spatiotemporal information in point cloud videos. P4Transformer and PST-Transformer~\cite{fan2021point} use spatiotemporal convolutions and Transformers to model temporal dynamics. PPTr~\cite{wen2022point} extends this with a multi-level Transformers, while LeaF~\cite{liu2023leaf} introduces an SE(3)-equivariant network to better disentangle motion information. Kinet~\cite{zhong2022no} proposes a two-stream framework that utilizes feature-level ST-surfaces in the dynamics learning branch to model dynamics. More recently, Mamba4D~\cite{liu2024mamba4d} presents a purely Mamba-based point cloud video backbone. 

While these models perform well in offline scenarios with access to entire video sequences, they are limited in practical settings where future frames are not available. For example, in real-time robotic perception, the network can only access historical information and cannot observe future frames. To address this, NSM4D~\cite{dong2023nsm4d} treats point cloud video understanding as an online real-time perception problem, using a neural scene model to record historical information. However, it reliance on a separate memory module increases system complexity and limits flexibility. In contrast, we propose a unified and lightweight network capable of handling both online real-time and offline settings, without requiring explicit scene memory mechanisms.

\noindent\textbf{Hybrid Mamba-Transformer Modules.} The combination of Mamba and Transformer into hybrid architectures has recently demonstrated impressive results across various domains. MambaVision~\cite{hatamizadeh2024mambavision} shows that adding a Transformer after Mamba greatly enhances its long-context modeling ability and scalability. Similarly, MaskMamba~\cite{chen2024maskmamba} proposes a hybrid Mamba-Transformer network for image generation, achieving strong performance. In the medical domain, HMT-UNet~\cite{zhang2024hmt} extends this idea to medical image analysis. SST~\cite{xusst} introduces the hybrid network to time series analysis, and PoinTramba~\cite{wang2024pointramba} demonstrates early success for point cloud analysis. In robotics, HMT-Grasp~\cite{xiong2024hmt} applies the hybrid design to grasping tasks, while MambaPolicy~\cite{cao2024mamba} proposes a hybrid network for efficient 3D Diffusion Policy learning. Longllava~\cite{wang2024longllava} extends hybrid architectures for Large Language Models. More Recently, MAP~\cite{liu2025map} introduces a pretraining framework tailored for hybrid Mamba-Transformer networks.  
MaST-Pre~\cite{shen2023masked} is a masked-based 4D point cloud video pretraining approach for the Transformer-based backbone. 


Our work addresses this gap by proposing a hybrid backbone (PointNet4D) tailored for point cloud video analysis, and introduces a pretraining strategy (4DMAP) to further enhance the hybrid architecture.



\section{PointNet4D: A Hybrid Backbone for Unified Online and Offline 4D Perception}

\subsection{The Need for a Unified 4D Backbone}

For 4D point cloud video understanding, most existing methods are designed for the offline setting, where the entire input sequence is available in advance. This allows models to access both past and future frames when processing each timestep, which is advantageous for post hoc analysis and offline applications. However, many practical applications, including robotics and AR/VR, require models to run under online real-time conditions, where the model must process data sequentially as it arrives. In this setting, future frames are unavailable, making online perception fundamentally more challenging due to the lack of bidirectional temporal context. Throughout this paper, we treat “online perception” and “online real-time perception” as equivalent terms.

To evaluate how well existing 4D backbones perform in real-time settings, we simulate both online and offline conditions a 4D action segmentation task using the HOI4D dataset~\cite{liu2022hoi4d}. This benchmark is closely related to robotic decision-making tasks, where agents must interpret temporal context from the past observations to make frame-wise predictions. Though the output differs: in robotics, it is actions, while in 4D action segmentation, it is action labels.

In our experiments, each method is given a point cloud video consisting of 150 frames and must predict the action category for each frame. In the offline setting, the network has access to the entire sequence, enabling the integration of bi-directional temporal information. In the online setting, only the current and past frames are available at inference time. All evaluated 4D backbones adopt a two-layer architecture: a single-frame feature extractor (either PointNet++~\cite{qi2017pointnet++} or Point4DConv~\cite{fan2021point}) followed by a temporal fusion module to obtain the final features—either a Transformer or Mamba. Based on the fusion module, we denote these variants as \textbf{P4Transformer}~\cite{fan2021point} and \textbf{P4Mamba}, respectively. Specifically, Point4DConv~\cite{fan2021point} uses three frames followed by P4Transformer. PointNet++~\cite{qi2017pointnet++} represents the direct use of single-frame information for prediction, while the Mambas/Transformers incorporate different temporal layers on top of the single-frame features from PointNet++. To ensure fair comparison in online conditions, P4Transformer~\cite{fan2021point} applies a causal attention mask, while P4Mamba uses unidirectional scanning; the offline versions use full attention or bidirectional scanning. In all offline settings, we select Point4DConv as the local feature extractor. For online settings, to ensure real-time performance and causality requirements, we choose PointNet++ as the extractor.

\begin{table}
    \centering
    \scriptsize
    \setlength{\tabcolsep}{4.5pt} 
    \renewcommand{\arraystretch}{1.1} 
    \begin{tabular}{c|c|ccccc}
    \hline
        {Offline} & Clips/s & Acc & Edit & F1@10 & F1@25 & F1@50\\
        \hline
        {Point4DConv~\cite{fan2021point}} & 19.9 &  51.2 & 29.5 & 34.1 & 27.2 & 17.4 \\
        \hline
        {P4Transformer~\cite{fan2021point}} & 7.0 & {71.2}  & {73.1} & {73.8} & {69.2} & {58.2} \\
        {P4Mamba} & {17.7} & 66.0 & 65.4 &69.0  & 63.3 & 50.8\\\rowcolor{cyan!10}
        {PointNet4D} &14.4 & 73.9  &73.8 & 75.3& 71.3& 61.3\\\rowcolor{cyan!10}
        {4DMAP} &14.4 & \textbf{77.2}  &\textbf{82.2} & \textbf{82.8}& \textbf{79.7} & \textbf{70.3} \\
        \hline
        \hline
        {Online}  & Clips/s & Acc & Edit & F1@10 & F1@25 & F1@50 \\
        \hline
        {PointNet++~\cite{qi2017pointnet++}} & 35.8 & 50.8  & 40.2 & 42.6 & 35.3 &23.5  \\
        \hline
        {P4Transformer~\cite{fan2021point}} & 26.6 & 66.7 & 62.0  & 65.3  & 59.8 & 46.3 \\
        {P4Mamba} & {33.4} & {71.0}  & {78.5} & {77.8} & {73.4} & {61.5} \\\rowcolor{cyan!10}
        {PointNet4D} & 31.8& 70.3 & 72.3&73.7 & 69.1& 57.2\\\rowcolor{cyan!10}
        {4DMAP} & \textbf{31.8} & \textbf{72.4}& \textbf{78.7}& \textbf{78.8}& \textbf{74.8}& \textbf{63.6}\\
    \hline
    \end{tabular}
    \caption{Performance Comparison of 4D Backbones in Online vs. Offline Settings on the HOI4D Action Segmentation Task. We compare our proposed PointNet4D and 4DMAP models with baseline architectures under both inference modes. Runtime (Clips/s) reflects real-time feasibility. Metrics include classification accuracy (Acc), edit score (Edit), and frame-wise F1 scores at three overlap thresholds (F1@10, F1@25, F1@50).}
    \vspace{-5mm}
    \label{tab:Pilot}
\end{table}

The experimental results are summarized in the Table~\ref{tab:Pilot}, with the following key observations: \textbf{First, the Transformer is more suitable for offline settings, while Mamba excels in online settings.} Specifically, P4Transformer outperforms P4Mamba in the offline case, while P4Mamba is superior in the online setting.
\textbf{Second, most of the computational cost in 4D backbone comes from the frame-wise feature extractor, not the temporal fusion layers.} For instance, architectures using Point4DConv run much slower (e.g., 7.0 clips/s for P4Transformer) than those based on PointNet++(26.6 clips/s). This shows that the Transformer and Mamba modules alone introduce minimal additional overhead.
\textbf{Third, PointNet++ paired with appropriate temporal fusion layers can yield strong performance at low cost.} 
We selected PointNet++ as our 3D feature extractor due to its balance of performance and computational efficiency.
P4Mamba in online settings achieves comparable results to P4Transformer in offline settings by relying on PointNet++ instead of Point4DConv, demonstrating its strong ability to model variable-length sequences. Moreover, P4Mamba provided a large performance boost to PointNet++ with less than 7\% additional computational overhead, demonstrating that temporal fusion layers play a key role even with lightweight backbones.

These insights motivate our design of \textbf{PointNet4D}, which combines the strengths of both P4Transformer and P4Mamba to achieve satisfactory performance in both offline and online settings. To keep computational demands low, we use PointNet++ as the feature extractor for single-frame point clouds and employ a hybrid Mamba-Transformer layer for temporal feature fusion. This architecture offers a unified solution that performs robustly in both online and offline 4D perception scenarios, while remaining efficient enough for real-time deployment.

\subsection{PointNet4D: A Universal 4D Backbone}

\textbf{Revisit PointNet++.} PointNet++~\cite{pointnet++} is an extension of PointNet~\cite{qi2017pointnet} that hierarchically applies PointNet to capture local features in a point cloud. Here is a basic outline of its components:    
Given a set of points \( \mathcal{P} = \{ \mathbf{p}_1, \mathbf{p}_2, \ldots, \mathbf{p}_N \} \), we perform sampling to select a subset of points \( \mathcal{P}_s \).  
For each sampled point \( \mathbf{p}_i \in \mathcal{P}_s \), a neighborhood is defined using a radius or k-nearest neighbors, forming a local region \( \mathcal{R}_i \).  
A mini-PointNet is applied to each local region \( \mathcal{R}_i \) to extract features:  

\begin{equation}
f(\mathcal{R}_i) = \gamma \left( \max_{\mathbf{p}_j \in \mathcal{R}_i} \left( \phi(\mathbf{p}_j) \right) \right)
\end{equation}
where \( \phi \) is a point-wise feature transformation, and \( \gamma \) is a symmetric aggregation function, typically max pooling.  

\noindent\textbf{From PointNet++ to PointNet4D.} 
Our proposed PointNet4D builds upon PointNet++ by incorporating temporal feature fusion using a hybrid Mamba-Transformer architecture. PointNet++ serves as the per-frame spatial feature extractor, while the hybrid temporal fusion module processes the sequence of extracted features to effectively handle spatiotemporal data. Given a point cloud sequence \( \mathcal{V} \) as  \( \{ \mathcal{P}_t \}_{t=1}^T   \) at each time step \( t \), PointNet++ extracts spatial features:  

\begin{equation}
\mathbf{F} = \text{PointNet++}(\mathcal{V}) 
\end{equation}
 
\noindent where \( \mathbf{F}_t \) represents the feature vector for the frame \( t \).   

To model temporal dynamics, our proposed PointNet4D introduces a hybrid Mamba-Transformer fusion layer that processes the sequence of extracted features \( \{ \mathbf{F}_t \} \):  
\begin{equation}
\mathbf{H} = \text{HybridLayer}(\mathbf{F}) 
\end{equation}

\noindent where each \(\textit{HybridLayer}\) is a combination of one Mamba layer with a transformer layer which combines the strengths of Mamba for online adaptability and Transformer for capturing long-range dependencies, and \(\mathbf{H}\) is the fused feature representation.   
  
The final output of PointNet4D is a temporally and spatially fused representation, which can be used for various downstream tasks:  
\begin{equation}
\mathbf{O} = \text{OutputLayer}(\mathbf{H}) 
\end{equation}
 
\noindent where \(\textit{OutputLayer}\) is a task-specific layer for predictions such as classification or segmentation. 

\begin{figure*}
    \centering
    \includegraphics[width=1\linewidth]{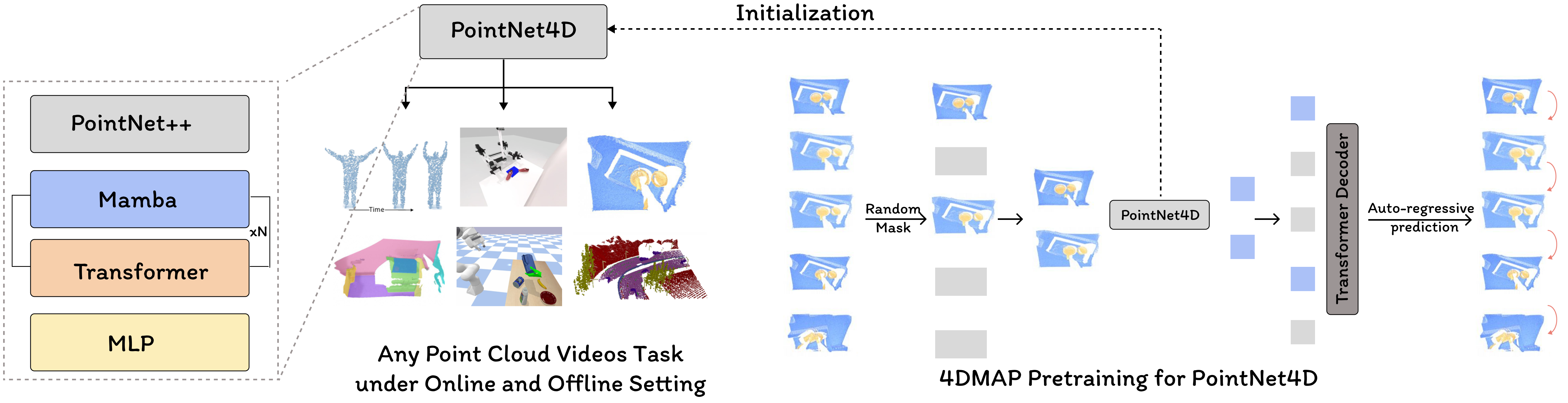}
    \caption{Method Overview. The left side of the diagram details the MTMT hybrid model's internal structure, a composite of PointNet++, Mamba, Transformer, and MLP designed for processing point cloud video tasks under both online and offline conditions. The right side outlines the 4DMAP pre-training strategy, which utilizes a frame-masking technique tailored for PointNet4D to maximize its performance potential across a broad spectrum of applications. This figure is adapted from ~\cite{shen2023masked,wen2022point,wang2024genh2r,mu2024robotwin,zhang2023complete}.}
    \label{fig:method}
    \vspace{-3mm}
\end{figure*}


\noindent\textbf{Evaluating Hybrid Fusion Strategies.}
We explore several strategies to combine Mamba (M) and Transformer (T) layers for temporal fusion and evaluate their impact on model accuracy. Specifically, we experiment with four layer orderings across two hybrid blocks: TTMM, MMTT, TMTM, and MTMT. As shown in Table~\ref{tab:Hybrid}, the MTMT configuration—alternating Mamba and Transformer layers—yields the highest accuracy. Thus, we adopt MTMT as the default fusion strategy in PointNet4D.

\begin{table}[t]
    \centering
    \small
    \begin{tabular}{c|cccc}
    \hline
       Hybrid  & TTMM & MMTT & TMTM &MTMT  \\
       \hline
       Accuracy & 69.1 & 69.6  &70.1 & \textbf{70.3}\\
    \hline
    \end{tabular}
    \vspace{-2mm}
    \caption{Effect of hybrid layers ordering on temporal fusion performance. Each configuration stacks two Mamba (M) and two Transformer (T) layers in different orders. MTMT yields the best performance and is used as our default fusion strategy.}
    \vspace{-3mm}
    \label{tab:Hybrid}
\end{table}

\noindent\textbf{Plug-and-Play Property.} PointNet4D is designed to be modular and extensible. Our temporal modeling components can be easily paired with other existing 3D powerful backbones. For example,  when we replace PointNet++ with a more powerful single-frame feature extractor, PointTransformer~\cite{fan2021point}, we refer to the resulting model as PointNet4D++. The only difference lies in the single-frame feature extractor, while the temporal fusion pipeline remains unchanged.


\section{4D Masked Autoregressive Vision Learners}
Pretraining is a widely adopted method to enhance model generalization and scalability. Masked Autoencoders (MAE)~\cite{he2022masked} have been proven especially effective in pre-training Transformers in the visual domain. Recently, MAP~\cite{liu2025map} proposed a pretraining method more suitable for the Mamba-Transformer backbone, which greatly boosts the potential of the hybrid framework by using a masked autoregressive approach. Building upon MAP, we introduce \textbf{4D} \textbf{M}asked \textbf{A}utoregressive \textbf{P}retraining (4DMAP), an extension designed for pretraining in temporal scenarios.

4DMAP differs from MAP in that it primarily handles temporal data and adopts a frame-masking strategy rather than random masking. Unlike the row-by-row decoding in image MAP, 4DMAP decodes on a per-frame basis, which aligns well with the temporal modeling capacity.
Compared to MAE, which reconstructs the entire 4D input at once, our 4DMAP decodes and autoregressively reconstructs one frame at a time, which is more beneficial to model temporal transitions.Compared to traditional autoregressive (AR) pretraining that decode one point or token at a time, 4DMAP reconstructs simultaneously within a single frame, which is more conducive to the network learning stronger spatial features. Therefore, 4DMAP can leverage the advantages of MAE in local spatial modeling and the advantages of AR in temporal correlation modeling, making it highly suitable for pretraining our proposed PointNet4D network. While this work focuses on point cloud videos, the principles of 4DMAP can be easily extended to RGB videos, offering potential for future applications.

\noindent\textbf{Overview.} As shown in the Figure~\ref{fig:method}, given a point cloud video, we first apply frame masking to it, then encode it using PointNet4D, and subsequently use a Transformer decoder to autoregressively reconstruct the original point cloud video on a frame-wise basis.

Consider a point cloud video as a sequence of 3D point clouds over time: 
\begin{equation}
\mathcal{V} = \{\mathcal{P}_t\}_{t=1}^T   
\end{equation}
  
We randomly select entire frames to mask. Let \(\mathcal{M} \subset \{1, 2, \ldots, T\}\) denote the indices of the masked frames.  
  
The goal is to predict the masked frames using visible frames. For each masked frame index \( t \in \mathcal{M} \), predict the frame based on the preceding visible frames: 
\begin{equation}
 p(\mathcal{P}_t \mid \mathcal{P}_{<t, \neg \mathcal{M}})  
\end{equation}
where \(\mathcal{P}_{<t, \neg \mathcal{M}}\) represents all preceding frames that are not masked.  
  
The loss function is the sum of the negative log-likelihoods of the predicted frames:  
\begin{equation}
 \mathcal{L} = -\sum_{t \in \mathcal{M}} \log p(\mathcal{P}_t \mid \mathcal{P}_{<t, \neg \mathcal{M}}) 
\end{equation}

Since we are pretraining PointNet4D, our method involves using PointNet4D to map the masked input into the feature space, followed by a Transformer decoder for frame-wise autoregressive decoding.


\noindent\textbf{Masking Strategy and Ratio.} 
The masking strategy and ratio play a critical role in mask-based pretraining frameworks. We conduct ablation studies to evaluate the impact of different masking strategies and ratios using the HOI4D action segmentation task in an online setting. As shown in Table~\ref{tab:strategies_ratio}, frame-level masking outperforms other strategies such as random or tube masking.
This suggests that in the MAP framework, where we employ autoregressive modeling, explicitly avoiding local copying in Transformers is unnecessary. Rather than focusing on features of locally grouped regions, the network benefits from concentrating on effectively capturing motion information across frames. Compared to Tube mask and random mask, frame mask not only allows PointNet4D to better model the motion information between frames but also enables it to better model the spatial information within frames. 
For the masking ratio, we find that a 50\% masking ratio to be optimal. This contrasts with findings of VideoMAE~\cite{tong2022videomae}, which advocate for higher masking ratios. However, in our experiments, a larger ratio here negatively impacts the network, as it makes predicting future frames from historical frames challenging, leading to model collapse. Thus, we selected the frame masking strategy and a 50\% masking ratio as our default settings.


\begin{table}[t]
    \small
    \centering
    \begin{tabular}{l|cccc}
        \multicolumn{5}{c}{\textbf{(a) Masking Strategy}} \\
        \hline
        Strategy & FS & Random & Tube & Frame \\
        \hline
        Accuracy & 70.3 & 71.2 & 71.5 & \textbf{72.4} \\
        \hline
    \end{tabular}
    \vspace{1mm}

    \begin{tabular}{l|ccccc}
        \multicolumn{6}{c}{\textbf{(b) Masking Ratio}} \\
        \hline
        Ratio & FS & 25\% & 50\% & 75\% & 90\% \\
        \hline
        Accuracy & 70.3 & 71.6 & \textbf{72.4} & 68.7 & 66.6 \\
        \hline
    \end{tabular}
    \caption{Impact of Masking Strategy and Ratio on 4DMAP Pretraining. (a) Frame-level masking and (b) a 50\% masking ratio yield the best performance. FS: Full Supervision (no pretraining).}
    \label{tab:strategies_ratio}
    \vspace{-2mm}
\end{table}


\noindent\textbf{4DMAP Transformer Decoder.}
After extracting features using PointNet4D, we use a Transformer Decoder to autoregressively reconstruct the original input on a frame-wise basis.
The Transformer decoder is preferred over Mamba because it can reconstruct frame-wise based on the encoder's features by applying a decoder mask. In contrast, the Mamba decoder, due to its unidirectional scanning nature, struggles to simultaneously reconstruct an entire local region. Compared to MAE-style decoders which reconstruct all frames simultaneously, our strategy better models the motion correlations between consecutive frames, facilitating the learning of motion information. Compared to AR, which autoregressively predicts one token at a time, our strategy better models the spatial information within frames, further enhancing the modeling of temporal information.

We performed ablation experiments to assess the impact of various decoding strategies on performance. Each decoding strategy determines the temporal window the network must process and interpret simultaneously. Table~\ref{tab:Decoder} shows that decoding fewer frames at once results in better performance. This confirms that larger windows increase task complexity and thus tend to reduce performance. By default, we decode each frame individually, treating each as a standalone decoding unit. Given a sequence length of 150 frames, this approach corresponds to a 0.7\% decoding ratio.
\begin{table}[t]
    \centering
    \small
    \setlength{\tabcolsep}{4pt} 
    \renewcommand{\arraystretch}{1.1} 
    \begin{tabular}{c|cccc}
    \hline
       Decoder window size & 1(0.7\%) & 10(7\%) & 25(17\%) & 50(33\%)  \\
       \hline
       Accuracy & \textbf{72.4} & 72.2  & 71.3 & 70.1\\
    \hline
    \end{tabular}
    \vspace{-3mm}
    \caption{Impact of decoding window size on accuracy. Smaller window sizes lead to better performance due to reduced complexity.}
    \label{tab:Decoder}
\end{table}

\noindent\textbf{Reconstruction Loss.}
Given the unordered nature of point clouds, we evaluate three loss functions commonly used in point cloud reconstruction:
Chamfer Distance (CD)-L2, CD-L1 and Earth Mover's Distance (EMD). As shown in Table~\ref{tab:Target}, CD-L2 provides a more faithful measure of the mean distance between the reconstructed and ground-truth point sets, thereby better preserving key features and the overall shape than other loss functions.
\begin{table}[t]
    \centering
    \small
    \begin{tabular}{c|cccc}
    \hline
       Loss Function & FS&CD-L2 & CD-L1 & EMD   \\
       \hline
       Accuracy & 70.3& \textbf{72.4} & 72.0  & 72.0 \\
    \hline
    \end{tabular}
    \vspace{-3mm}
    \caption{Impact of loss function. }
    \vspace{-3mm}
    \label{tab:Target}
\end{table}

\section{Experiments}
\label{sec:experiments}
In this section, we first validate the effectiveness of PointNet4D on the online 4D point cloud video understanding task. Additionally, we further demonstrate that our PointNet4D is also effective for an offline setting. The same datasets as the downstream tasks are used for all pre-training.

\subsection{Online 4D Point Cloud Video Understanding}
\textbf{Setting.} 
The HOI4D dataset~\cite{liu2022hoi4d} contains 3,863 point cloud sequences, each with 150 frames, collected from 9 participants interacting with 800 distinct object instances, annotated across 16 categories within 610 indoor environments. 
For action segmentation, we use the official split, comprising 2,971 training scenes and 892 test scenes, with frame-level action labels for 19 unique classes. For semantic segmentation, each point is labeled across 39 classes. Implementation details can be found in the Supplementary Material.

\subsubsection{Online HOI4D Action segmentation Task}
\textbf{Results.}
Table~\ref{table:hoi4das_online} shows the substantial advantages of our PointNet4D series over previous methods. Compared to PointNet++, adding a lightweight hybrid temporal fusion layer nearly doubles performance, underscoring the value of temporal fusion modeling.
Our method also outperforms NSM4D~\cite{dong2023nsm4d}, even though NSM4D uses four times more frames (600 vs. our 150), demonstrating PointNet4D's efficiency.
Examining the impact of pretraining, we observe that 4DMAP yields considerable performance gains over PointNet4D. The improvement is particularly pronounced in Edit and F1 scores, indicating that pretraining enhances not only the network’s accuracy but also the smoothness and consistency of its predictions. Furthermore, the performance boost from PointNet4D++ over PointNet4D reaffirms the effectiveness of our approach and demonstrates the model's compatibility with other 3D perception backbones, showcasing the strong potential for broad application.


\begin{table}[t]
\setlength{\tabcolsep}{0.15mm}
    \scriptsize
\scriptsize{
\begin{center}
\label{table:hoi4das}
\newcolumntype{Y}{>{\centering\arraybackslash}X}
{
\begin{tabularx}{\columnwidth}{>{\centering} m{0.3\columnwidth}|>{\centering} m{0.12\columnwidth}|>{\centering} m{0.125\columnwidth}|Y|Y|Y|Y}
\hline\noalign{\smallskip}
\textbf{Online Setting} & Length   & Acc & Edit & F1@10 & F1@25 & F1@50\\
\hline
{PointNet++~\cite{qi2017pointnet++}} & 1 & 50.8  & 40.2 & 42.6 & 35.3 &23.5  \\
{P4Transformer~\cite{fan2021point}} & 150 & 66.7 & 62.0  & 65.3  & 59.8 & 46.3 \\
{P4Mamba} & 150 & 71.0  & 78.5 & 77.8& 73.4 & 61.5 \\
{PPTr~\cite{wen2022point}} & 150 & 69.7  & 64.5 &69.4 & 64.8 & 52.9 \\
NSM4D~\cite{dong2023nsm4d}& 150& 67.8& 63.2& 68.0& 63.7& 51.9\\
NSM4D~\cite{dong2023nsm4d}& 600 &71.3 &68.0 &72.1 &68.1 &56.5\\
\hline\rowcolor{cyan!10}
{PointNet4D} & 150& 70.3 & 72.3&73.7 & 69.1& 57.2\\\rowcolor{cyan!10}
{4DMAP} & 150 & 72.4& 78.7& 78.8& 74.8& 63.6\\\rowcolor{cyan!10}
{PointNet4D++} & 150& 76.1 &72.9 &76.4&72.9 &63.2 \\\rowcolor{cyan!10}
{4DMAP++} & 150 & \textbf{77.7}& \textbf{80.6} &\textbf{82.6} &\textbf{79.6} &\textbf{69.9} \\
\hline
\end{tabularx}
}
\vspace{-3mm}
\caption{Online action segmentation on HOI4D dataset. PointNet4D and its variants (marked in blue) outperform prior works, especially with 4DMAP pretraining.}
\label{table:hoi4das_online}
\end{center}
}
\end{table}

\subsubsection{Online HOI4D Semantic Segmentation Task}
\textbf{Results.} Table~\ref{table:hoi4dss_online} indicates that our PointNet4D variants not only excels in frame-level tasks but also achieves superior performance in more detailed semantic segmentation tasks. 
Our method surpasses existing methods even with a 3-frame time window, and extending the window to 10 frames leads to further improvements, showcasing the adaptability of our network in capturing finer-grained temporal and spatial information.


\begin{table}[t]
\setlength{\tabcolsep}{0.15mm}
    \scriptsize
\footnotesize{
\begin{center}
\label{table:hoi4dss}
\newcolumntype{Y}{>{\centering\arraybackslash}X}
{
\begin{tabularx}{\columnwidth}{>{\centering} m{0.47\columnwidth}|>{\centering} m{0.2\columnwidth}|Y}
\hline\noalign{\smallskip}
\textbf{Online Setting} & Clip Length   & mIoU\\
\hline
PointNet++\cite{qi2017pointnet++} & 1 & 26.6\\
P4Transformer\cite{fan2021point} & 3	 &  41.6\\
P4Mamba & 3	 &  39.8\\
PPTr\cite{wen2022point} & 3 & 42.7 \\
\hline\rowcolor{cyan!10}
PointNet4D & 3 &  42.6\\\rowcolor{cyan!10}
PointNet4D++ & 3 &  42.8\\\rowcolor{cyan!10}
PointNet4D++ & 10 & 43.2 \\\rowcolor{cyan!10}
4DMAP & 3 & 43.9 \\\rowcolor{cyan!10}
4DMAP++ & 3 & 44.9 \\\rowcolor{cyan!10}
4DMAP++ & 10 & \textbf{45.3} \\
\hline

\end{tabularx}
}
\vspace{-3mm}
\caption{Online semantic segmentation on HOI4D dataset. mIoU improves as we incorporate 4DMAP and stronger backbones.}
\label{table:hoi4dss_online}
\vspace{-8mm}
\end{center}
}
\end{table}

\subsection{Offline 4D Point Cloud Video Understanding}

\textbf{Setting.}
In the offline setting, the entire point cloud sequence is available at inference.
We compared our approach with other methods on the HOI4D dataset~\cite{liu2022hoi4d}, validating its superior performance. We further evaluated our network on several other datasets including MSRAction-3D~\cite{li2010action}, NTU-RGBD~\cite{shahroudy2016ntu}, SHREC'17~\cite{de2017shrec}, SHREC'17~\cite{de2017shrec} and  Synthia 4D~\cite{ros2016synthia}. Detailed introductions to each dataset and more offline task results can be found in the supplementary material. For each dataset, we employed a 5-layer hybrid architecture and applied 4DMAP pretraining, reporting results accordingly. We also enhanced the network by using PPTr as the encoder, creating PointNet4D++, which incorporates multi-level spatial information awareness. The corresponding pretrained network is designated as 4DMAP++.

\begin{table}[t]
\setlength{\tabcolsep}{0.15mm}
    \scriptsize
\scriptsize{
\begin{center}
\label{table:hoi4das}
\newcolumntype{Y}{>{\centering\arraybackslash}X}
{
\begin{tabularx}{\columnwidth}{>{\centering} m{0.3\columnwidth}|>{\centering} m{0.12\columnwidth}|>{\centering} m{0.125\columnwidth}|Y|Y|Y|Y}
\hline\noalign{\smallskip}
\textbf{Offline Setting} & Length   & Acc & Edit & F1@10 & F1@25 & F1@50\\
\hline
\multicolumn{7}{c}{\textbf{Supervised Training}} \\
\hline
Point4DConv~\cite{fan2021point}& 3  &  51.2 & 29.5 & 34.1 & 27.2 & 17.4 \\
P4Transformer~\cite{fan2021point} & 150 & 71.2 & 73.1 & 73.8 & 69.2& 58.2\\
PPTr~\cite{wen2022point} & 150 &  77.4 & 80.1 & 81.7 & 78.5 & 69.5\\
LeaF~\cite{liu2023leaf} &150 & {79.4}  & {83.9}  & {85.0} & {81.9} & {73.3}\\ 
\hline\rowcolor{cyan!10}
PointNet4D &150 & 73.9  &73.8 & 75.3& 71.3& 61.3\\\rowcolor{cyan!10}
PointNet4D++ &150 &78.5  &80.7 &81.6 &78.7 & 69.6\\
\hline
\multicolumn{7}{c}{\textbf{Pretraining}} \\
\hline
PPTr+STRL~\cite{huang2021spatio} &150& 78.4& 79.1& 81.8 &78.6 &69.7\\
PPTr+VideoMAE~\cite{tong2022videomae} &150& 78.6& 80.2& 81.9& 78.7& 69.9\\
P4T+C2P~\cite{zhang2023complete} &150& 73.5& 76.8 &77.2& 72.9& 62.4\\
PPTr+C2P~\cite{zhang2023complete} &150& 81.1& 84.0& 85.4& 82.5& 74.1\\
\hline\rowcolor{cyan!10}
4DMAP&150  & 77.2  &82.2 & 82.8& 79.7 & 70.3 \\\rowcolor{cyan!10}
4DMAP++&150  & 81.5  &84.5 & 85.6& 82.5&74.5  \\

\hline
\end{tabularx}
}
\vspace{-3mm}
\caption{Offline action segmentation on HOI4D dataset.}
\vspace{-3mm}
\label{table:hoi4das_offline}
\end{center}
}
\end{table}

\begin{figure*}[!h]
    \centering
    \includegraphics[width=\linewidth]{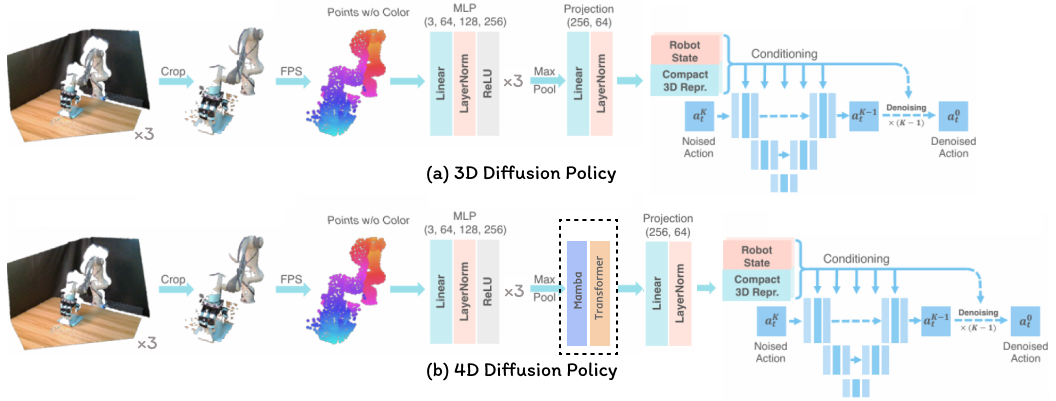}
    \vspace{-6mm}
    \caption{The comparison between the 3D Diffusion Policy (DP3)~\cite{ze20243d} and the 4D Diffusion Policy (DP4). }
    \vspace{-3mm}
    \label{fig:dp4}
\end{figure*}

\subsubsection{Offline HOI4D Action Segmentation Task}
\textbf{Results.} Table~\ref{table:hoi4das_offline} shows that, in a supervised training setup, our PointNet4D series networks consistently deliver notable performance gains. Unlike LeaF~\cite{liu2023leaf}, which increases computational overhead with SE(3)-Equivariant networks and a dual-branch structure, our network maintains a lightweight design while achieving comparable or superior performance. In the pretraining evaluations, our approach also demonstrates steady improvements, reaching state-of-the-art outcomes. These findings confirm that our method is highly effective not only in the online setting but also brings substantial benefits in the offline setting.


\subsubsection{Offline HOI4D Semantic Segmentation Task}
\textbf{Results.} 
Table~\ref{table:hoi4dss_offline} shows that in the offline semantic segmentation task, our approach yields obvious improvements. After pretraining, both 4DMAP and 4DMAP++ surpass the performance of models pretrained on CrossVideo with multimodal data. This highlights the strong compatibility of our 4DMAP pretraining strategy with the PointNet4D network, underscoring its effectiveness even without relying on multimodal pertaining.


\begin{table}[t]
\setlength{\tabcolsep}{0.15mm}
    \scriptsize
\footnotesize{
\begin{center}
\label{table:hoi4dss}
\newcolumntype{Y}{>{\centering\arraybackslash}X}
{
\begin{tabularx}{\columnwidth}{>{\centering} m{0.47\columnwidth}|>{\centering} m{0.2\columnwidth}|Y}
\hline\noalign{\smallskip}
\textbf{Offline Setting} & Clip Length   & mIoU\\
\hline
\multicolumn{3}{c}{\textbf{Supervised Training}} \\
\hline
P4Transformer\cite{fan2021point} & 3	 & 40.1 \\
PPTr\cite{wen2022point} & 3 & 41.0 \\
LeaF\cite{liu2023leaf} & 3 &  {43.5}\\
\hline\rowcolor{cyan!10}
PointNet4D & 3 & 41.6 \\\rowcolor{cyan!10}
PointNet4D++ & 3 & 41.9 \\
\hline
\multicolumn{3}{c}{\textbf{Pretraining}} \\
\hline
P4Transformer+C2P~\cite{zhang2023complete} &3 &41.4\\
PPTr+STRL~\cite{huang2021spatio} &3& 41.2\\
PPTr+VideoMAE~\cite{tong2022videomae}& 3 &41.3\\
PPTr+C2P~\cite{zhang2023complete} &3 &42.3\\
\hline\rowcolor{cyan!10}
4DMAP & 3 &  42.7\\\rowcolor{cyan!10}
4DMAP++ & 3 & 45.0 \\
\hline
\end{tabularx}
}
\vspace{-3mm}
\caption{Offline semantic segmentation on HOI4D dataset.}
\label{table:hoi4dss_offline}
\vspace{-8mm}
\end{center}
}
\end{table}

\section{Application for Robotics}
In this section, we demonstrate how our proposed PointNet4D enhances robotic learning performance when applied to two distinct downstream frameworks. 


\begin{table*}[h]
    \centering
    \scriptsize
    \begin{tabular}{l|ccccccc}
        \midrule
        Task & Apple Cabinet Storage & Pick Apple Messy & Shoe Place & Empty Cup Place & Mug Hanging & Shoes Place & Diverse Bottles Pick \\
        \midrule
        DP~\cite{chi2023diffusion} & 72 & - & 9 & 20 & 0 & 0 & 0 \\
        DP3~\cite{ze20243d} & 61 & 7 & 42 & 71 & 14 & 8 & 37 \\\rowcolor{cyan!10}
        DP4 & 93(+32) & 8(+1) & 50(+8) & 71 & 29(+15) & 8 & 39(+2) \\\rowcolor{cyan!10}
        DP4* & 95(+35) & 10(+3) & 52(+10) & 72(+1) & 30(+16) & 9(+1) & 42(+5) \\
         \midrule
        Task & Dual Bottles Pick\tiny{(Hard)}	 & Dual Bottles Pick\tiny{(Easy)} & Container Place &Block Hammer Beat  & Block Handover & Blocks Stack\tiny{(Easy)} & Blocks Stack\tiny{(Hard)} \\
         \midrule
        DP~\cite{chi2023diffusion} & 28 & 54 & 0 & 0 & 28 & 2 & 0 \\
        DP3~\cite{ze20243d} & 48 & 74 & 70 & 50 & 68 & 14 & 1 \\\rowcolor{cyan!10}
        DP4 & 48 & 80(+6) & 71(+1) & 54(+4) & 94(+26) & 23(+9) & 1 \\\rowcolor{cyan!10}
        DP4* & 50(+2) & 83(+9) & 72(+2) & 57(+7) & 96(+28) & 26(+12) & 2(+1) \\
         \midrule
    \end{tabular}
    \caption{Comparison of DP, DP3, and our DP4 on various manipulation tasks in RoboTwin~\cite{mu2024robotwin} benchmark. Our DP4 model achieved notable performance gains compared to DP3~\cite{ze20243d}, which utilizes PointNet++, with only an additional 0.9M parameters(the default parameters for DP3 is around 260M) and minimal computational overhead. Specifically, our DP4 improved success rates by 32\%, 26\%, and 15\% in the Apple Cabinet Storage, Block Handover, and Mug Hanging tasks, respectively.}
    \label{tab:performance_comparison}
\end{table*}

\begin{table*}[h]
\centering
\small
\begin{tabular}{c|ccc|ccc}
\hline
\multirow{2}{*}{Train on DexYCB~\cite{chao2021dexycb}} & \multicolumn{3}{c}{Sequential} & \multicolumn{3}{c}{Simultaneous} \\ \cline{2-7} 
                              & Success Rate     & Time    & Average Success   & Success Rate     & Time    & Average Success    \\ \hline
OMG Planner~\cite{wang2019manipulation}       & 62.50 & 8.31 & 22.5 & -     & -    & -    \\ \hline
GA-DDPG~\cite{wang2022goal}             & 50.00 & \textbf{7.14} & 22.5 & 36.81 & \textbf{4.66} & 23.6 \\ \hline
Handover-Sim2real~\cite{christen2023learning}   & \textbf{75.23} & 7.74 & \textbf{30.4} & 68.75 & 6.23 & 35.8 \\ \hline
GenH2R~\cite{wang2024genh2r}          & - & - & - & 74.31 & 6.91 & 34.81 \\ \hline\rowcolor{cyan!10}
4DIL               & - & - & - & \textbf{78.47} & 6.31 & \textbf{40.40} \\ \hline
\end{tabular}
        \vspace{-3mm}
    \caption{Comparison of our proposed 4DIL with previous methods on the HandoverSim~\cite{chao2022handoversim} benchmark. Our 4DIL achieves the best success rate in the more challenging simultaneous setting.}
    \label{tab:performance_comparison_handover}
    \vspace{-2mm}
\end{table*}


\subsection{4D Diffusion Policy (DP4)}
\textbf{Setting.}
We first explore whether PointNet4D can improve the performance of the existing Diffusion Policy frameworks~\cite{chi2023diffusion, ze20243d,ke20243d}. 
All experiments are conducted on the RoboTwin benchmark~\cite{mu2024robotwin}, using 
the same code repository and set the head camera's field of view to 37°. 
The 3D Diffusion Policy demonstrates leading advantages in data efficiency and performance, prompting us to replace its PointNet++ visual encoder with PointNet4D to form 4D Diffusion Policy (DP4), as illustrated in Figure~\ref{fig:dp4}. 
To ensure fairness, we statistically analyze the results under identical settings across all configurations, using a time window of 3. All other settings adhere to RoboTwin's default configuration, ensuring a fair comparison of all results.

%

\noindent\textbf{Results.}
The results in Table~\ref{tab:performance_comparison} show that with only an additional 0.9M parameters, DP4 with PointNet4D brings obvious overall improvements. Specifically, the success rate for Apple Cabinet Storage increased by 32\%, for Mug Hanging by 15\%, and for Block Handover by 26\%. DP4* represents the performance when using the pretrained weights from HOI4D for network initialization. As observed, pretraining with human data through 4DMAP consistently enhances the robot's performance on downstream tasks. This not only validates the effectiveness of 4DMAP but also underscores the vast potential of utilizing human 4D data to pretrain robot perception networks.


\subsection{4D Imitation Learning (4DIL)}
\textbf{Setting.} 
We further validated the potential of PointNet4D within an imitation learning framework, using the HandoverSim~\cite{chao2022handoversim} benchmark for evaluation. All settings were kept consistent with the default configuration of HandoverSim. In the sequential setting, the robot performs the grasping action after the human hand has remained stationary. This scenario does not require 4D information, as both the hand and object states remain unchanged once stationary, relying solely on 3D perception. In contrast, the simultaneous setting involves both the human hand and the object in motion during the robot’s grasping action, making this setup much more challenging than the sequential one. We reproduced the results of GenH2R in the Simultaneous setting and confirmed them with the authors. To ensure a fair comparison, we used the same codebase and replaced PointNet++ with PointNet4D in the GenH2R model, utilizing a 3-frame time window. We report the results of our method in the more challenging simultaneous setting.

\noindent\textbf{Results.} 
The results in Table~\ref{tab:performance_comparison_handover} demonstrate with only 0.9M additional parameters, we achieved significant overall improvements in the simultaneous setting. From the Handover-Sim2Real results, it is evident that the simultaneous setting, due to its higher difficulty, leads to significantly lower success rates. Our method outperforms existing approaches in both the sequential and the more challenging simultaneous settings. This result highlights that our PointNet4D seamlessly integrates into the imitation learning framework, showcasing its strong performance and versatility.


\section{Conclusion}
\label{sec:conclusion}
We present \textbf{PointNet4D}, a lightweight yet powerful backbone designed for online and offline 4D point cloud video understanding. To further enhance its capability, we introduce \textbf{4DMAP}, a tailored masked autoregressive pretraining strategy that improves temporal modeling and generalization. Built upon PointNet4D, we develop two practical robotic learning systems—\textbf{4D Diffusion Policy (DP4)} and \textbf{4D Imitation Learning (4DIL)}—demonstrating PointNet4D’s plug-and-play adaptability across tasks. 
Our extensive experiments across 9 tasks and 2 major robotic benchmarks show that PointNet4D and its derivatives consistently outperform prior existing methods, while maintaining low computational overhead. Our work highlights the value of unified online/offline 4D perception and paves the way for real-time robotic applications that rely on dynamic 3D understanding.

{\small
\bibliographystyle{unsrt}
\bibliographystyle{ieeenat_fullname}
\bibliography{11_references}
}

\clearpage \appendix 
\clearpage
\setcounter{page}{1}
\maketitlesupplementary

\section*{Implementation Details}
In our approach, we adopted PointNet++ followed by five hybrid temporal fusion layers as the standard backbone for PointNet4D. In the action segmentation task, the network processes variable-length inputs per time step, from 1 up to 150 frames. In semantic segmentation, the network inputs the current frame along with the previous two frames to output dense semantic predictions for the current frame. For pretraining PointNet4D, we employed a 4-layer Transformer Decoder with a masked autoregressive approach for frame reconstruction. The pretrained encoder then initializes PointNet4D and is fine-tuned for downstream tasks. 
As mentioned in the method section, the only difference between PointNet4D++ and PointNet is that PointNet4D++ uses PointTransformer\cite{wen2022point} as the feature extractor for per-frame point clouds. The pretraining strategy corresponding to PointNet4D++ is referred to as 4DMAP++. All experiments were conducted on 8 A800 GPUs. Pretraining was performed over 300 epochs, followed by 80 epochs of fine-tuning. Using a batch size of 32, we adopted a learning rate of 0.03. The other settings are consistent with P4Transformer\cite{p4d} and PPTr\cite{wen2022point}.

\section*{Datasets Details for Other Offline Tasks}
MSRAction-3D\cite{li2010action}: This dataset contains 567 videos across 20 daily action categories, with each video averaging around 40 frames. We followed the standard setup, using 270 videos for training and 297 for testing. We use the 24 frames as default. NTU-RGBD\cite{shahroudy2016ntu}: Comprising 56,880 videos in 60 fine-grained action categories, video lengths range from 30 to 300 frames. In the cross-subject setting, we split the dataset into 40,320 training and 16,560 testing videos. SHREC'17\cite{de2017shrec}: This dataset includes 2800 videos across 28 gesture classes. NvGesture\cite{molchanov2016online}: Composed of 1532 videos covering 25 gesture classes, with 1050 videos for training and 482 for testing. Synthia 4D\cite{ros2016synthia}: A synthetic outdoor driving dataset, Synthia4D generates 3D videos based on the Synthia dataset, capturing six driving scenarios with moving objects and cameras. We followed previous work, splitting the dataset into 19,888 training, 815 validation, and 1,886 testing frames. 

\section*{Additional Results on Offline Tasks}
\textbf{3D Action Recognition on NTU-RGBD.} We further validated the effectiveness of our approach for offline tasks on additional datasets. On NTU-RGBD, our PointNet4D achieved state-of-the-art results under supervised training and comparable results to M2PSC after pretraining with 4DMAP. Notably, M2PSC utilized human pose tracking for pretraining, whereas our pretraining did not incorporate any human priors. Beyond human action classification, our method also demonstrated competitive performance in gesture recognition tasks. These experiments further support the generality and effectiveness of our approach.

\begin{table}[h]
    \centering
    \scriptsize

    \setlength{\tabcolsep}{1.6mm}
    \begin{tabular}{lc}
    \toprule
    \textbf{Methods}    & \textbf{Acc.} \\
    \midrule
    3DV-Motion~\cite{3dv}                    & 84.5 \\
    3DV-PointNet++~\cite{3dv}                & 88.8 \\
    PSTNet~\cite{pstnet}                     & 90.5 \\
    PSTNet++~\cite{fan2021deep}              & 91.4 \\
    Kinet~\cite{zhong2022no}                 & 92.3 \\
    P4Transformer~\cite{p4d}                 & 90.2 \\
    PST-Transformer~\cite{fan2022point}      & 91.0 \\ \rowcolor{cyan!10}
    PointNet4D~\cite{p4d}                 &90.5  \\
    \midrule
    PSTNet + PointCPSC\cite{sheng2023point} (50\% Semi-supervised) &88.0\\
    PSTNet + PointCMP\cite{shen2023pointcmp} (50\% Semi-supervised) &88.5\\
    PSTNet + CPR\cite{sheng2023contrastive} (End-to-end Fine-tuning) &91.0\\
    P4Transformer + MaST-Pre\cite{shen2023masked} (50\% Semi-supervised)& 87.8\\
    P4Transformer + MaST-Pre\cite{shen2023masked} (End-to-end Fine-tuning) &90.8\\
    P4Transformer + M2PSC\cite{hanmasked} (50\% Semi-supervised)& 88.7\\
    P4Transformer + M2PSC\cite{hanmasked} (End-to-end Fine-tuning) &91.3\\ \rowcolor{cyan!10}
    4DMAP (50\% Semi-supervised)& 88.8\\ \rowcolor{cyan!10}
    4DMAP (End-to-end Fine-tuning) &90.9\\
    \bottomrule
    \end{tabular}
    \label{NTU}
    \caption{Action recognition accuracy (\%) on NTU-RGBD.}
\end{table}

\noindent\textbf{Offline 4D Action Recognition Tasks on MSRAction3D, SHREC’17 and NvGesture.} We also conducted experiments on the 4D action recognition task on MSRAction3D. PointNet4D-PST refers to our model, which is built upon the PST-Transformer and incorporates our hybrid Mamba-Transformer layer. While M2PSC is a framework specifically designed for human point cloud video analysis, utilizing trajectory tracking of human points for pretraining, our method, as a general-purpose architecture, has already demonstrated competitive performance when compared to specialized methods tailored to the human domain. In addition, we have validated the effectiveness of PointNet4D and 4DMAP on the SHREC’17~\cite{de2017shrec} and NvGesture~\cite{molchanov2016online} gesture recognition tasks.

\begin{table}[t]
    \centering
    \scriptsize
    \setlength{\tabcolsep}{1.3mm}
    \begin{tabular}{l|l|c}
    \toprule
    \multicolumn{2}{c|}{\textbf{Methods}} & \textbf{Accuracy (\%)}\\
    \midrule
    \multirow{9}{*}{{\centering{Supervised Learning}}} &
    MeteorNet~\cite{MeteorNet}    & 88.50 \\
    & PSTNet~\cite{pstnet}        & 91.20 \\
    & PSTNet++~\cite{fan2021deep} & 92.68 \\  
    & Kinet~\cite{zhong2022no}    & 93.27 \\
    & PPTr~\cite{wen2022point}    & 92.33 \\
    & P4Transformer~\cite{p4d}             & 90.94 \\
    & PST-Transformer~\cite{fan2022point}  &93.73 \\
      & Mamba4D  &92.68 \\
    & \cellcolor{cyan!10}PointNet4D &\cellcolor{cyan!10}91.61 \\
    & \cellcolor{cyan!10}PointNet4D-PST &\cellcolor{cyan!10}93.75 \\
    \midrule
    \multirow{9}{*}{\centering{End-to-end Fine-tuning}} 
    & PSTNet + PointCPSC~\cite{sheng2023point}& 92.68\\
    & PSTNet + CPR~\cite{sheng2023contrastive}& 93.03\\
    & PSTNet + PointCMP~\cite{shen2023pointcmp}&93.27\\
    & P4Transformer + MaST-Pre~\cite{shen2023masked}&91.29\\
    & PST-Transformer + MaST-Pre~\cite{shen2023masked}&94.08\\
    & P4Transformer + M2PSC~\cite{hanmasked}  & 93.03 \\
    & PST-Transformer + M2PSC~\cite{hanmasked}  &94.84 \\
    & \cellcolor{cyan!10}4DMAP &\cellcolor{cyan!10}92.65 \\
    & \cellcolor{cyan!10}4DMAP-PST &\cellcolor{cyan!10}94.76 \\
    \bottomrule
    \end{tabular}
    \caption{Action recognition accuracy on MSRAction-3D.}
    \label{MSRAction-3D}
\end{table}

\begin{table}[h]
    \centering
    \scriptsize

    \begin{tabular}{lcc}
    \toprule
    \textbf{Methods} &\textbf{NvG} &\textbf{SHR}\\
    \midrule
    FlickerNet~\cite{flickernet}           & 86.3    & -    \\
    PLSTM~\cite{min2020efficient}     & 85.9    & 87.6 \\
    PLSTM-PSS~\cite{min2020efficient}      & 87.3    & 93.1 \\
    Kinet~\cite{zhong2022no}               & 89.1    & 95.2 \\
    P4Transformer~\cite{p4d} \ (30 Epochs)   & 84.8   & 87.5\\
    P4Transformer~\cite{p4d} \ (50 Epochs)   & 87.7   & 91.2 \\\rowcolor{cyan!10}
    PointNet4D\ (30 Epochs) &85.0& 87.7\\\rowcolor{cyan!10}
    PointNet4D \ (50 Epochs) & 87.7 &91.0\\
    \midrule
    P4Transformer + MaST-Pre\cite{shen2023masked} \ (30 Epochs)  & 87.6 & 90.2 \\
    P4Transformer + MaST-Pre\cite{shen2023masked} \ (50 Epochs)  & 89.3 & 92.4 \\
    P4Transformer + M2PSC\cite{hanmasked} \ (30 Epochs) &88.0& 90.9\\
    P4Transformer + M2PSC\cite{hanmasked} \ (50 Epochs) &89.6 &92.8\\\rowcolor{cyan!10}
    4DMAP \ (30 Epochs) & 87.8& 90.5\\\rowcolor{cyan!10}
    4DMAP \ (50 Epochs) & 89.6&92.6\\
    \bottomrule
    \end{tabular}
    \caption{Gesture recognition accuracy (\%) on NvG and SHR.}
    \label{NvGesture}
\end{table}

\noindent\textbf{Offline Semantic Segmentation on Synthia 4D.} We conducted experiments on an outdoor autonomous driving dataset to validate the effectiveness of our method. Our approach continues to yield significant performance improvements on this dataset, underscoring its versatility across various domains and data formats. This demonstrates the potential of our method as a robust general-purpose 4D backbone.

\newcommand{\tabincell}[2]{\begin{tabular}{@{}#1@{}}#2\end{tabular}} 
\begin{table}[h]
\setlength{\tabcolsep}{0.2mm}
\tiny{
\begin{center}

\label{table:synthia4d}
\begin{tabular}{l|c|cccccccccccc|c}
\hline\noalign{\smallskip}
Offline Setting & Clip Length & Bldn& Road& Sdwlk & Fence& Vegittn & Pole & Car & T.Sign & Pedstrn & Bicycl & Lane & T.Light& mIoU\\
\noalign{\smallskip}
\hline
\noalign{\smallskip}
3D MinkNet14~\cite{choy20194d}&1 & 89.39& 97.68&69.43 &86.52 &98.11 &97.26 &93.50 &79.45 &92.27 & 0.00 & 44.61 & 66.69 & 76.24\\
4D MinkNet14~\cite{choy20194d}& 3 & 90.13& 98.26& 73.47& 87.19& 99.10& 97.50& 94.01& 79.04& {92.62}& 0.00 & 50.01 & 68.14 & 77.24\\
\hline
\noalign{\smallskip}
PointNet++~\cite{pointnet++} & 1& 96.88 &  97.72 & 86.20 & 92.75 & 97.12 & 97.09 & 90.85 & 66.87 & 78.64 & 0.00 & 72.93 & 75.17 & 79.35 \\
MeteorNet-m~\cite{choy20194d}& 2 & {98.22} & 97.79 & 90.98 & 93.18 & 98.31 & 97.45 & 94.30 & 76.35  & 81.05 & 0.00 & 74.09 & 75.92& 81.47 \\
MeteorNet-l~\cite{choy20194d} & 3 & 98.10 & 97.72  & 88.65 & 94.00 & 97.98 & 97.65 & 93.83 & {84.07}  & 80.90  &0.00 & 71.14 & 77.60 & 81.80\\
\noalign{\smallskip}
\hline
\noalign{\smallskip}
P4Transformer~\cite{fan2021point} & 1 & 96.76 & 98.23 & 92.11 & 95.23 & 98.62 & 97.77 & 95.46 & 80.75 & 85.48 & 0.00 & 74.28 & 74.22 & 82.41\\
P4Transformer~\cite{fan2021point} & 3 & 96.73 & 98.35 & 94.03 & 95.23 & 98.28 & 98.01 & 95.60 & 81.54 & 85.18 & 0.00 & 75.95 & {79.07} & 83.16 \\
\noalign{\smallskip}
\hline\rowcolor{cyan!10}
PointNet4D & 3 & 97.23 & 98.35 & 94.73 & 95.93 & 99.08 & 98.11 & 97.60 & 81.04 & 87.18 & 0.00 & 76.90 & 77.97 & 83.67 \\\rowcolor{cyan!10}
4DMAP & 3 &97.30 &98.31 & 95.13& 96.79& 99.35& 98.16& 97.91& 80.88&89.60&0.00&78.01&77.69&84.10\\
\hline
\end{tabular}
\caption{Evaluation for semantic segmentation on Synthia 4D.}
\end{center}
}
\end{table}

\section*{More Details of 4D Imitation Learning(4DIL)}
We also validated the significant potential of PointNet4D within an imitation learning framework, using the HandoverSim\cite{chao2022handoversim} benchmark for evaluation. All settings were kept consistent with the default configuration of HandoverSim. The Sequential setting represents the scenario where the robot performs the grasping action after the human hand has remained stationary. This setup does not require 4D information, as the hand and object states do not change once they are stationary, relying solely on 3D perception. The Simultaneous setting, on the other hand, involves both the human hand and the object in motion during the robot’s grasping action, making this setup significantly more challenging than the Sequential one.  We reproduced the results of GenH2R in the Simultaneous setting and confirmed them with the authors. To ensure a fair comparison, we used the same codebase and replaced PointNet++ with PointNet4D in the GenH2R model, using a 3-frame time window. We report the results of our method in the more challenging Simultaneous setting.
\begin{figure}
    \centering
    \includegraphics[width=1\linewidth]{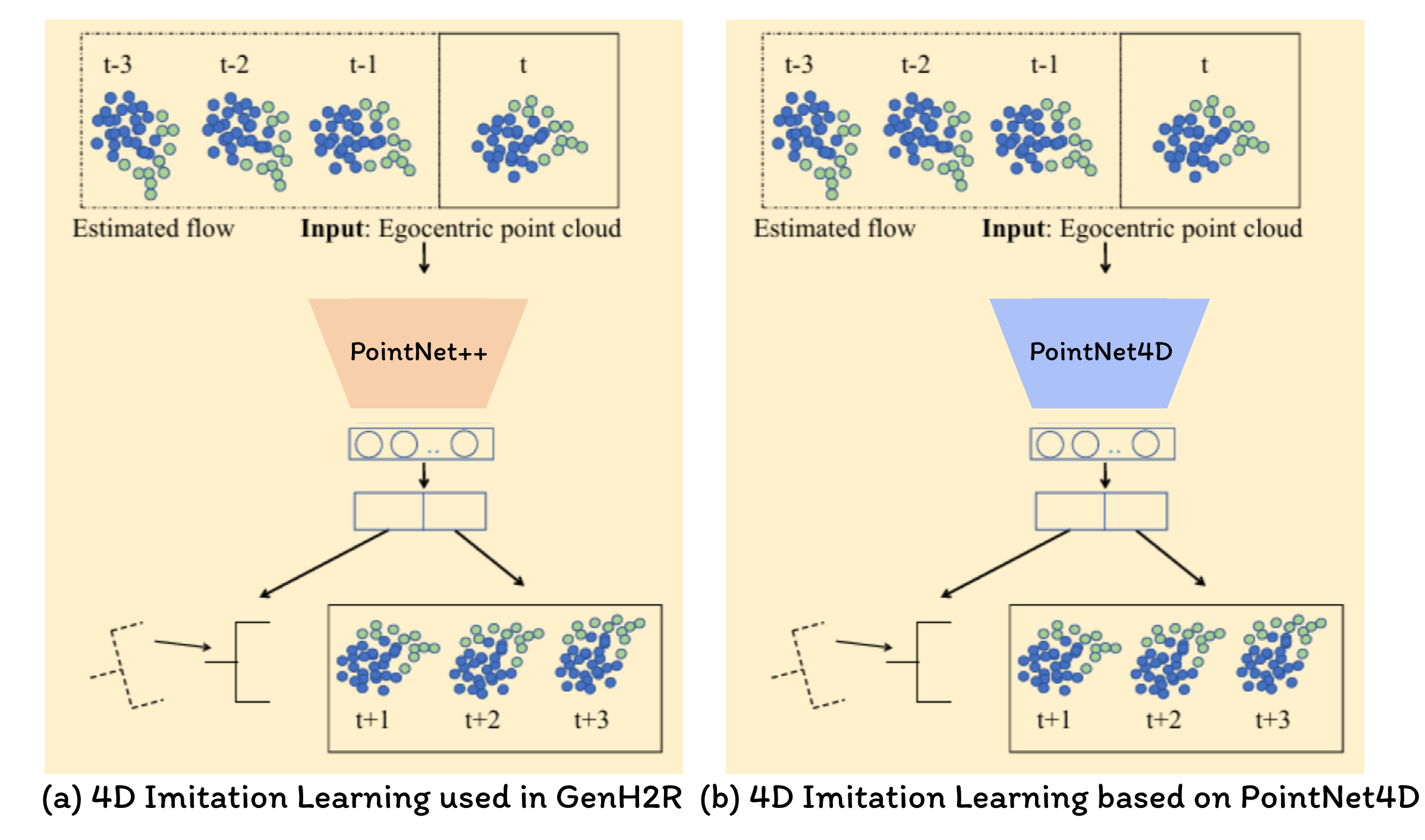}
    \vspace{-6mm}
    \caption{Comparison of our 4DIL with GenH2R~\cite{wang2024genh2r}.}
    \vspace{-5mm}
    \label{fig:4dil}
\end{figure}


\end{document}